\documentclass[3p,authoryear,preprint]{elsarticle}

\def\ie{\textit{i.e.} }
\def\eg{\textit{e.g.} }

\def\RRCNN{R\textsuperscript{2}CNN}

\usepackage{amsmath}
\usepackage[mathscr]{euscript}	

\usepackage{siunitx}

\usepackage{graphicx}
\usepackage{epsfig}
\usepackage{epstopdf}
\usepackage{subfig}

\usepackage{tabularx}
\usepackage	{multirow}			
\usepackage	{multicol}

\usepackage[applemac]{inputenc}
\usepackage[T1]{fontenc}

\usepackage{moreverb}

\usepackage	[colorlinks=true,linkcolor=black,citecolor= blue,breaklinks]{hyperref}

\usepackage{textgreek}  

\usepackage{amsmath, amssymb}
\usepackage{import} %
\usepackage[]{siunitx} 
\usepackage{bm}  %
\newcommand\gommettes{0.16\textwidth}

\def\graphpath{./}
\graphicspath{\graphpath}

\usepackage{placeins}
\let\Oldsection\section
\renewcommand{\section}{\FloatBarrier\Oldsection}
\makeatletter
\def\ps@pprintTitle{%
   \let\@oddhead\@empty
   \let\@evenhead\@empty
   \def\@oddfoot{\reset@font\hfil\thepage\hfil}
   \let\@evenfoot\@oddfoot
}
\makeatother
\begin{document}
\begin{frontmatter}
\date{}

\title{\textbf{Technical Report}\\ Towards \textit{DeepSpray}: Using Convolutional Neural Network to post-process Shadowgraphy Images of Liquid Atomization}
\author[aff1]{G.~Chaussonnet}
\ead{geoffroy.chaussonnet@kit.edu}
\author[aff1]{C.~Lieber}
\author[aff1]{Y.~Yikang}
\author[aff1]{W.~Gu}
\author[aff2]{A.~Bartschat}
\author[aff2]{M.~Reischl}
\author[aff1]{R.~Koch}
\author[aff2]{R.~Mikut}
\author[aff1]{H.-J.~Bauer}
\address[aff1]{Karlsruher Institut f{\"u}r Technologie - Institut f{\"u}r Thermische Str{\"o}mungsmaschinen, Karlsruhe, Germany}
\address[aff2]{Karlsruher Institut f{\"u}r Technolgie - Institut f{\"u}r Automation und angewandte Informatik, Eggenstein-Leopoldshafen, Germany}

\begin{abstract}
This technical report investigates the potential of Convolutional Neural Networks to post-process images from primary atomization. Three tasks are investigated. First, the detection and segmentation of liquid droplets in degraded optical conditions. Second, the detection of overlapping ellipses and the prediction of their geometrical characteristics. This task corresponds to extrapolate the hidden contour of an ellipse with reduced visual information. Third, several features of the liquid surface during primary breakup (ligaments, bags, rims) are manually annotated on 15 experimental images. The detector is trained on this minimal database using simple data augmentation and then applied to other images from numerical simulation and from other experiment.\\
In these three tasks, models from the literature based on Convolutional Neural Networks showed very promising results, thus demonstrating the high potential of Deep Learning to post-process liquid atomization. The next step is to embed these models into a unified framework \textit{DeepSpray}.\\\\
\textit{NB: The resolution of the figures was decreased to fulfill ArXiv's requirement. For a full resolution, please refer to:
\href{https://doi.org/10.5445/IR/1000097897/v3}{https://doi.org/10.5445/IR/1000097897/v3}}
\end{abstract}
\end{frontmatter}

\section{Introduction \label{sec_intro}}

\begin{figure}[!htb]
	\centering
		\includegraphics[width=0.28\textwidth,keepaspectratio]{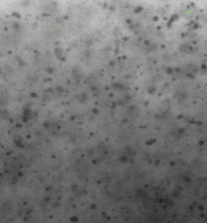}
		\includegraphics[width=0.32\textwidth,keepaspectratio]{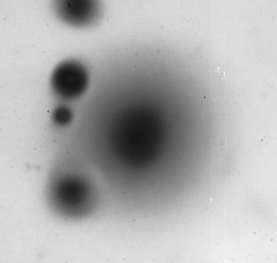}
		\includegraphics[width=0.36\textwidth,keepaspectratio]{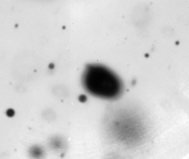}
		\caption{Examples of snapshots were accuracy is limited. Right: heterogeneous background. Center: defocused droplet. Right: large deformation spot}
		\label{fig_intro_example}
\end{figure}

In the context of post-processing experiment of liquid atomization and sprays, there are different techniques, each owing their own advantages. Interferometric techniques such as Laser Diffraction Technique (LDT) or Phase Doppler Anemometry (PDA) are the standard methods to extract the characteristics of a spray. LDT relies on the Mie diffraction and provides the droplet size distribution in the form of a histogram. Thus, the spray droplets are considered as a statistical set. It requires very little calibration and is often commercialized as a ready-to-use instrument. On the other hand, PDA relies on the refraction of light inside each individual droplet, and therefore is able to account each single droplet passing through the control volume, giving access to the diameter and velocity of the same droplet. PDA requires the careful alignment of pairs of laser beams, which require more overheads than LDT. A common denominator of LDT and PDA is that they can measure only near-to-spherical droplets, which limits the minimum distance between the nozzle and the measurement volume.
When the droplets are not spherical, the PDA technique usually reject the droplet, which is trackable whereas LDT leads to a deviation \citep{dumouchel2014laser}. This requirement of spherical droplets severely limits the utility of LDT and PDA for investigating the early stages of a liquid spray, \ie the primary atomization, in which liquid emerging from the nozzle is disrupted into non-spherical liquid structures such as ligaments, bags and blobs.\\
In the last two decades, a technique was developed based on shadowgraphy, or back-light illumination. This consists in illuminating the liquid from the background and photographing/recording the scene from the front with a CCD camera. The liquid appears in shade in  white background. A contour detection algorithm is used for segmentation of the structures and extract their surface, which can be extrapolated to equivalent diameters.
The advantage of this technique is that it gives a deeper and more tangible insight of the primary breakup process, as all structures are captured. As it is based on optical images, the structures need to be in the focal plane of the imaging system. The counterpart is that defocused structures or droplets appear blurry (Fig.~\ref{fig_intro_example} center, right), which increases the uncertainty of the measurements. Also, the space resolution depends on the lens/objectives and on the size of CCD chip.
The traditional techniques to detect droplets include contour detection based on pixel threshold \citep{doi:10.1002/ppsc.200300897, doi:10.1002/ppsc.200400898}, or used in combination with a second thresholding based on a wavelet transform \citep{blaisot2005droplet}. Other popular method to detect liquid structures are based on a combination of the extraction of the intensity gradients (\eg use of the Canny filter \citep{jeong2007investigation} or Sobel filter \citep{kanopoulos1988design}) and the use of Hough transform to discriminate the most-probable object. This has shown good results for overlapping object \citep{yuen1990comparative}. However, after applying the filter, a threshold is still necessary to discriminate the foreground from the background, which is influenced by the background homogeneity.
Recently, \citet{lieber2019experiments} succeeded to refine the resolution to 1 \textmu m/pixel with the use of a long distance microscope. 
A technique used at Institut f{\"u}r Thermische Str{\"o}mungsmaschinen (ITS) to reduce the shortcoming of defocused droplets is to use calibration images prior to the experiment. Calibrating plates, made of opaque circles of known diameter, are photographed a several offset distances from the focal plane. Thus this step mimics the recording of defocused spray droplets. As the opaque circles are defocused, their contour is blurry, which leads to a gradient of intensity on the droplet contour. By correlating the intensity gradient to the distance to focal plane and to the droplet diameter, it is possible to obtain a deterministic correlation map between measured and real droplet diameter. More details can be found in \citep{warncke2017experimental}. 
However, despite this correcting method, there are still caveats that diminish the applicability of the method. The most important problem is when the background is not homogeneous or when there is a gradient of luminosity (Fig.~\ref{fig_intro_example} left). This occurs in case of large liquid volume ratios (dense spray), where the liquid out of the focal plane absorbs a significant amount of the backlight.
In this case, the aforementioned traditional algorithms to detect droplets cannot accurately detect the contour of liquid structure.
Indeed, droplet detection can also be found in other fields such as life science with droplet microfluidics \citep{guo2012droplet,grosche2019microfluidic}, additive manufacturing \citep{dugas2005droplet} or environment \citep{kanthan2015rain}. However, in these applications, the detection is somewhat favored by almost spherical droplets with a homogeneous background, which is not the case in images from primary breakup experiments. 
In addition to heterogeneous background, when large defocused droplets are passing between the objective and the focal plane, this leads to large deformation spots that distort the measurement (Fig.~\ref{fig_intro_example} right).
In addition contour detection algorithm are fundamentally limited because they can extract a contour only, but they cannot extract information based on the pixel value inside the contour. This severely limits the quantities of information to exploit from shadowgraphy images.
The texture of the complex structures such as ligament, blobs or bags carry valuable information of the shape and volume distribution of the liquid to be atomized.
Recently, liquid atomization was investigated by means of X-ray \citep{kastengren2017measurements,halls2017high,machicoane2019synchrotron}. This technique allows to scan the liquid volume during breakup and it provides a 2D picture with different slices superimposed on each other. Therefore droplets, ligaments, blobs and bags are superimposed on the same image. This increases the amount of information to be extracted, which leads to more complicated post-processing tools. To the authors knowledge, there is no such tools able to separate the different structures from an X-ray images in the context of liquid breakup.
As an intermediate conclusion, shadowgraphy and X-ray imaging offer a deep insight of the primary breakup process, but there is a lack of tools to extract physical information from the snapshots. Techniques based on mathematical analysis of the contour or the shape of liquid structures perform poorly because of the distortion of the liquid surface, which is intrinsic to the liquid during atomization. 
\\
In the last decade, there was a tremendous breakthrough of Machine-Learning (ML) methods in different aspects of data treatment such as speech recognition, text analysis, language translation, image classification, feature detection, etc. The basic concept of machine learning is to train an algorithm with some data already labeled or sorted (supervised training). During the training, the algorithm elaborates hierarchical models which are not defined by the user, 
in order to reduce the deviation between the output and the target.
Among the different families of ML algorithms, one of the most successful is Deep Learning (DL) \citep{lecun2015deep}. It consists in processing the data through a chain of layers made of numerical neurons which act as logical gates, and whose coefficients are fine-tuned automatically during the training phase. The framework for DL is the concept of Artificial Neural Networks (ANN).
Originally, in ANNs, all neurons of one layer are connected to every neuron of the next layer. This is called a fully connected layer. With at least one coefficient per connection, this leads to an enormous number of coefficients that would prohibits the training and use of deep neural network, even on nowadays supercomputers. For instance, an image of 28x28 resolution leads to 784 input pixels. With 5 subsequent layers of 50 neurons, this would lead to 245 billion connections, \ie at least 245 billion parameters to train.
The solution to this caveat is to correlate neighboring pixels by a convolution, thus reducing the output number of one layer. This allows to decrease the dimensionality of the data and increase the depth of the neural network. This technique, called Convolutional Neural Network (CNN),
mimics the chain of image treatment between eyes and brain in biology.
CNNs demonstrated their efficiency by almost halving the error rate in image classification \citep{krizhevsky2012imagenet}. Since then, they are used in rapidly evolving area where the treatment of complex scenery is critical (autonomous driving, medicine).\\

In this technical report, CNNs are applied to post-process snapshots of experiments related to liquid atomization. The results are compiled from the master thesis of \cite{yikang2019application} and \cite{gu2019application}, and they constitute a proof-of-concepts of the use CNN to post-process experiment/simulation of liquid atomization. 
Three questions are tackled:
\begin{enumerate}
\item Can CNNs be used to improve the estimation of droplet diameter from shadowgraphy images in degraded conditions?
\item Are CNNs able to the extrapolate geometric characteristics of objects that overlap each other?
\item Can CNNs sort the different distorted liquid structures that are characteristic features of primary breakup?
\end{enumerate}
Preliminary answers to these questions are given in Section \ref{dense_spray}, \ref{sec_overlapping_ellipses} and \ref{sec_prefilmer}, respectively.
\section{CNN models and indicators of relevance \label{methods_and_co}}

\subsection{CNN models}

The different structures of neural nets are presented in the following. Each of them will be used in for different tasks. Depending of the task, (i) image segmentation to estimate the droplet diameters or (ii) object detection to extract the characteristics structure of primary breakup, different type of neural networks will be used.

\subsubsection{UNET \label{sssec_UNET_pres}}

\begin{figure}[h]
	\centering
		\includegraphics[width=0.55\textwidth,keepaspectratio]{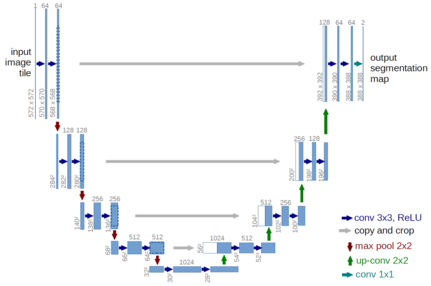}
		\caption{Architecture of UNET networks, from \cite{ronneberger2015u}.}
		\label{fig_UNET}
\end{figure}

The UNET architecture belongs to the category of autoencoder algorithm whose global pipeline is depicted in Fig.~\ref{fig_UNET}. It is made of an encoder and a decoder. In the encoder step, a Fully Convolutional Neural Network (FCNN) is classically used to reduce the dimensionality of the data and extract the features in the latent space. In the decoder step, the information of the latent space is decoded by the reversing the convolutional process of the encoder. This is done by replacing pooling operators by upsampling operators. Thus, it re-increases the dimensionality of the data, up to the last layer, which has the same resolution as the input. The decoder is almost the mirror of the encoder, which can be any type of CNN: AlexNet \citep{krizhevsky2012imagenet}, GoogleNet \citep{szegedy2015going}, etc.\\
\subsubsection{SSD: Single Shot Detector \label{sssec_SSD_pres}}

\begin{figure}[h]
	\centering
		\includegraphics[width=0.85\textwidth,keepaspectratio]{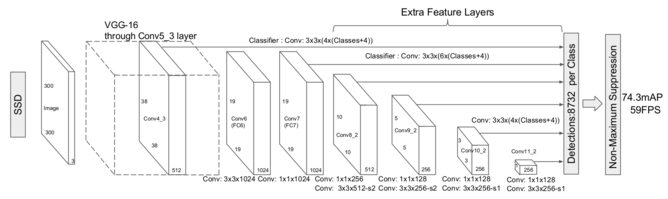}
		\caption{Architecture of the SSD model, from \citet{liu2016ssd}.}
		\label{fig_SSDD_Model}
\end{figure}

In their original paper, \cite{liu2016ssd} proposed their approach as a solution for an embedded system to provide a fast detection. Therefore the authors aimed at reducing the complexity of detection algorithm by proposing a sequential, continuous and unified framework, avoid any loops or querying from different processing tools. This explains the name 'single-shot'. The principle is to split the image on a background grid and produces a collection of bounding boxes linked to each cell of the grid. Then, the presence of objects of a given feature (or class) is evaluated by a score. To increase the resolution of small object, several background grids of larger resolution are used sequentially. The final detection is made by taking into account the detected objects on all scales. To achieve this task,  \cite{liu2016ssd} used the VGG16 net from \cite{simonyan2014very}. The model is depicted in Fig.~\ref{fig_SSDD_Model}.

\subsubsection{YOLO: You Only Look Once \label{sssec_YOLO_pres}}

\begin{figure}[h]
	\centering
		\includegraphics[width=0.75\textwidth,keepaspectratio]{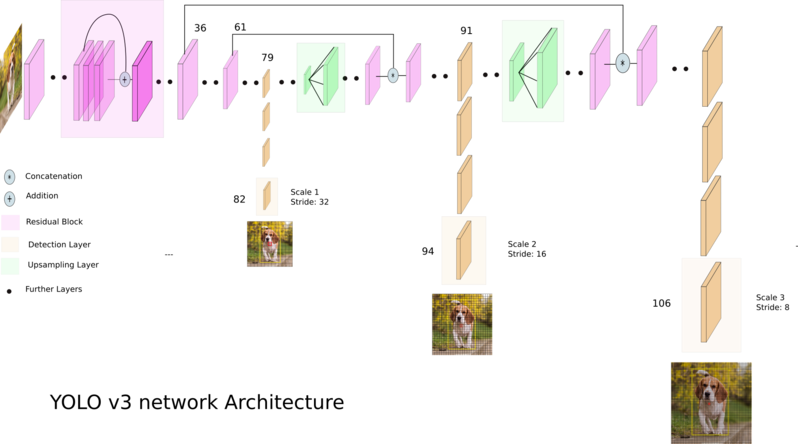}
		\caption{Architecture of YOLO v3, from \citet{kathuria2018what}.}
		\label{fig_YOLOv3}
\end{figure}

\cite{redmon2016you} proposed a model for object detection in same the philosophy as SSD: a simplified sequential architecture that can provide object detection at a high FPS, hence its name 'You Only Look Once'. In the original publication \citep{redmon2016you}, YOLO had only one scale of resolution, but after successive improvements \citep{redmon2017yolo9000,redmon2018yolov3}, three levels of resolution are provided. In the last release, the feature extractor of YOLO is a CNN embedding 53 convolutional layers baptized 'Darknet53'. Figure~\ref{fig_YOLOv3} illustrates the architecture of YOLOv3.

\subsubsection{RRCNN \label{sssec_RRCNN_pres}}
\begin{figure}[h]
	\centering
		\includegraphics[width=\textwidth,keepaspectratio]{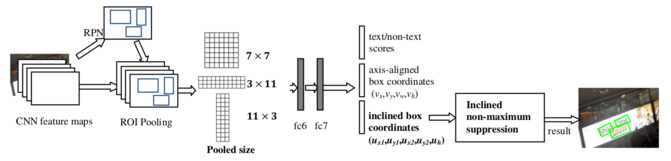}
		\caption{Architecture of R\textsuperscript{2}CNN, from \citep{jiang2017r2cnn}}
		\label{fig_RRCNN}
\end{figure}

The Rotational Region CNN (RRCNN) from \citet{jiang2017r2cnn} is based on a Faster R-CNN \citep{ren2015faster} and was proposed for detection of text oriented in various directions.
R-CNN stands for Region-based CNN. In this model, there is a decoupling between the tasks of (i) feature detection and (ii) bounding box proposal. Theses tasks are then achieved by two different models, a feature detector network and a Region Proposal Network (RPN) that exchange information. The decoupling of these task can lead to dramatic loss in performance. To circumvent this weakness, \cite{ren2015faster} proposed to share common layers between the two networks. Thus, the RPN benefits from the classifications operated by the feature detector. The interested reader is referred to \citep{ren2015faster} for more details. The principle of the RRCNN model is depicted in Fig.~\ref{fig_RRCNN}. The model uses the information from the feature maps and from the RPL to perform a Region Of Interest (ROI) Pooling. Then the pooled regions are injected into two full connected layers (fc6 and fc7) to propose (i) scores, (ii) axis-aligned boxed and (iii) inclined boxes. An inclined Non-Maximum Suppression is applied to select the final boxes. Note that contrary to SSD and YOLO, \RRCNN\ allow rotates the bounding boxes, so that they better align with the detected object.

\subsection{Indicators of relevance \label{ssec_scores}}

In the following, the performances will be quantitatively assessed in terms of Recall, Precision and mean Averaged Precision (mAP):
\begin{equation}
\text{Recall} = \frac{\text{TP}}{\text{TP} + \text{FN}}
\quad \text{,} \quad
\text{Precision} = \frac{\text{TP}}{\text{TP} + \text{FP}}
\end{equation}
where TP, FN and FP stand for True Positive (object detected correctly), False Negative (object missed) and False Positive (object detected whereas there was nothing to be detected).\\
The Recall is the ratio of objects correctly detected over all objects to detect. The Precision is the ratio of objects correctly detected over all detections.
The mean Average Precision is obtained by taking the mean value of the Precision for all images having a Recall in a given class (\eg 0.3 $<$ Recall $<$ 0.4). The mAP is then the average of all Precisions averages on all classes of Recall. In other words, mAP flattens the weight to calculate the mean Precision over the range of Recall, in order to avoid an over-contribution of a given class of Recall.\\
Also, the Intersection over Union (IoU) will be used to estimate how precise the location of the bounding box is. The IoU consists in computing the intersection of the bounding boxes given by the Ground Truth and by the prediction. The surface area of the intersection is computed and normalized by the surface area of the union of the two bounding boxes. For a perfect match, the intersection coincides with the union, leading to an IoU of 1. The IoU decreases as the precision decreases.

\section{Extraction of droplet diameter in degraded conditions \label{dense_spray}}

One caveat with contour detection using pixel threshold is the accuracy of the diameter. Usually, the pixel value used for the threshold is arbitrary set as fraction of the median of the pixel values (typically 80\%), providing an acceptable results. However in case of gradient of background luminosity, the measured diameter of the same droplet depends on the position of the droplet in the image, which is not acceptable. There is a possibility to split the image in smaller images where the background would be more homogeneous \citep{lieber2019experiments}, but this still leads to a deviation of the estimated droplet diameter. In this section, a method based on a CNN will be tested to circumvent these caveats.

\begin{figure}[h]
	\centering
		\includegraphics[width=0.495\textwidth,keepaspectratio]{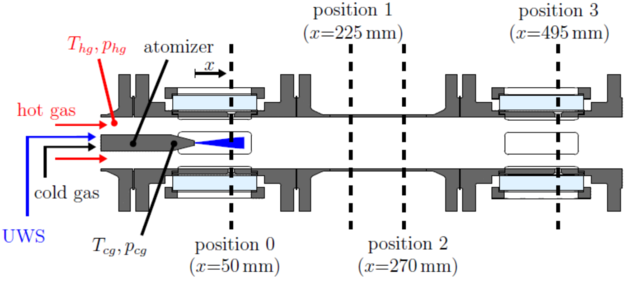}
		\includegraphics[width=0.4\textwidth,keepaspectratio]{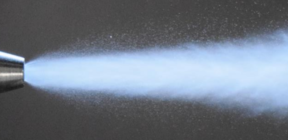}
		\caption{Left: Sketch of the test-rig, from \cite{lieber2019experiments}. Right: Close-up of the primary dense spray.}
		\label{fig_testrig_chris}
\end{figure}

The images to be processed are taken from an experiment made at ITS in the context of exhaust gas after treatment using urea-water injection at high ambient temperature and pressure. The test-rig, depicted in Fig.~\ref{fig_testrig_chris} (left) consists of a nozzle discharging into a hot pipe flow. Several optical accesses are mounted along the axial direction to assess the spray quality versus the distance to the nozzle. A solution of urea-water mixture is turned into a spray using a effervescent atomizer.
In this type of nozzle, the gas is mixed to the liquid inside an inner mixing chamber. The resulting emulsion then exits the atomizer where primary breakup occurs.
In order to avoid recirculation zones and to regulate the temperature, hot gas is injected around the injector. A snapshot of the spray in its primary phase is given in Fig.~\ref{fig_testrig_chris} (right). Close to the nozzle the spray is very dense and due to evaporation the temperature is not homogeneous. These reasons forbid the use of PDA or LDT, so that shadowgraphy imaging is used. In order to capture volumetric characteristics of the spray, several slices are measured by moving the focal plane of the imaging system in the transverse direction, from z=0 mm (center line) to z = -5 mm. The depth of field is 0.4 mm and the dimension of the image are 1.2 mm by 0.9 mm, leading to a resolution of 0.75 \textmu m / pixel. More details can be found in \citep{lieber2019experiments}.

\subsection{Selected architecture and workflow}

Obviously, even with data augmentation, the number of elements available in the training pool is limited. Therefore it is necessary to use a method that can be trained with limited labeled images. The UNET architecture by \cite{ronneberger2015u} presented in Section~\ref{sssec_UNET_pres} was selected for this aspect.\\%
In our case, the input is the raw image from the experiment, and the output is the segmented image where each pixel belonging to a droplet is marked, as illustrated in Fig.~\ref{fig_sample_training_data_base}. Then applying a classical contour detection can isolate every cluster and compute their equivalent diameter.\\
In the present study two architectures were tested: VGG16 by \cite{simonyan2014very} and ResNet by \cite{he2016deep}. It was found that VGG performs better, \ie it detects more droplets compared to the ResNet architecture.
In order to increase the resolution of the tool, the image is split in smallest images with a resolution of 256\textsuperscript{2}. This increases the detection rate for small droplets.
\subsection{Generating the database}

\begin{figure}[h]
	\centering
		\includegraphics[width=0.85\textwidth,keepaspectratio]{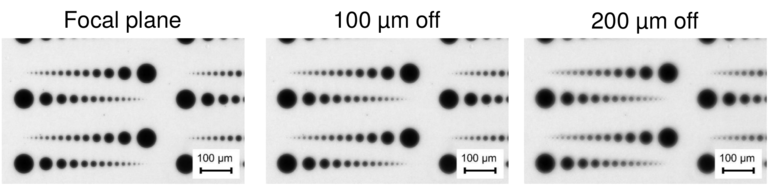}
		\caption{Calibration image on the focal plane (left), at 100 \textmu m (center) and 200 \textmu m (right) beyond the FP.}
		\label{fig_calibration}
\end{figure}

\begin{figure}[h]
	\centering
\hfill
		\includegraphics[width=0.25\textwidth,keepaspectratio]{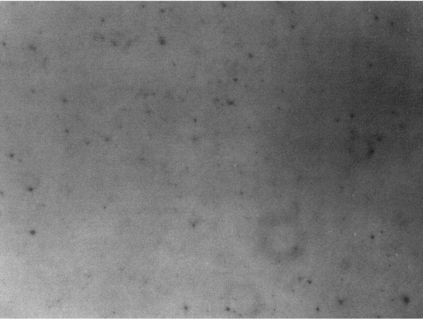}
\hfill
		\includegraphics[width=0.25\textwidth,keepaspectratio]{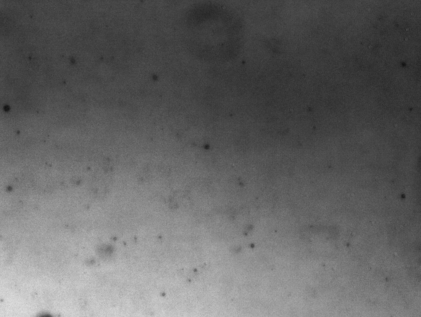}
\hfill
		\includegraphics[width=0.25\textwidth,keepaspectratio]{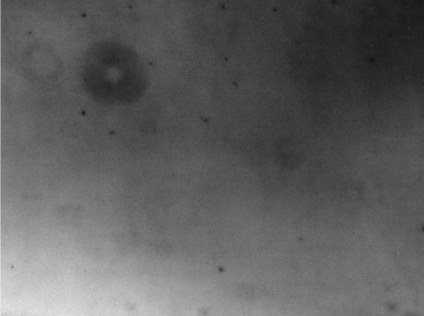}
\hfill
		\caption{Realistic backgrounds}
		\label{fig_realistic_bsckgrounds}
\end{figure}

The training database is created as follows. To calibrate the standard post-processing tool, images of a calibration plate are taken with different offset from the focal plane (Fig.~\ref{fig_calibration}). Each circle is isolated for all images to constitute a pool of isolated droplets, whose diameter is known. Then, the droplets are randomly chosen from the pool to be inserted into an artificial image, that will be used to train the CNN. Since the diameter of the circle is known, the ground truth, \ie the learning material, is straightforward to generate. In order to increase the training database, basic data augmentation techniques are used such as horizontal/vertical flip, contrast reduction, noise, heterogeneous background and geometric deformation. In addition, in order to make the CNN more robust, realistic backgrounds are taken from experimental image (Fig.~\ref{fig_realistic_bsckgrounds}) and are populated with droplets drawn from the pool. These images of real background are first cleaned from droplets, \ie droplets are erased from the background. This leads to white circles that looks unrealistic, but that does not affect the training of the CNN.

\begin{figure}[h]
	\centering
		\hfill
		\includegraphics[width=\gommettes,keepaspectratio]{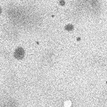}
		\includegraphics[width=\gommettes,keepaspectratio]{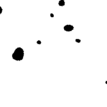}
		\hfill
		\includegraphics[width=\gommettes,keepaspectratio]{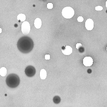}
		\includegraphics[width=\gommettes,keepaspectratio]{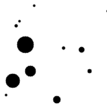}
		\hfill
		\includegraphics[width=\gommettes,keepaspectratio]{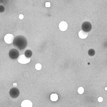}
		\includegraphics[width=\gommettes,keepaspectratio]{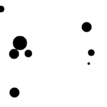}
		\\
		\hfill
		\includegraphics[width=\gommettes,keepaspectratio]{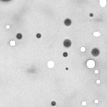}
		\includegraphics[width=\gommettes,keepaspectratio]{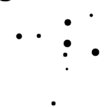}
		\hfill
		\includegraphics[width=\gommettes,keepaspectratio]{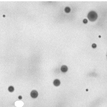}
		\includegraphics[width=\gommettes,keepaspectratio]{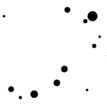}
		\hfill
		\includegraphics[width=\gommettes,keepaspectratio]{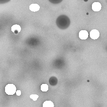}
		\includegraphics[width=\gommettes,keepaspectratio]{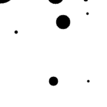}
		\caption{Example of training images: each pair represent the input image and the ground truth.}
		\label{fig_sample_training_data_base}
\end{figure}

Finally, the training database is made from 5000 artificial images generated by the opaque circles from the calibration images superimposed on real heterogeneous backgrounds. 
The training parameters are 8 images/batch, 512 batches/epoch and 5 epochs.
The corresponding ground truth is easily generated, accordingly to the actual diameter of the calibration black circles. Some samples of the training images are depicted in Fig.~\ref{fig_sample_training_data_base}. One can see the white spots that come from the cleaning of the real background images.
The content of the pool of real circles is summarized in Table~\ref{tab_elem_pool}.

\begin{table}[!htb]
	\centering
	\caption{Number of elements in the training pool.}
	\begin{tabular}{l c c}
		\hline
		\hline
		Diameter & [ \textmu m] & 60, 40, 30, 25, 20, 18\\
		  & & 16, 14, 12, 10, 8, 6, 4, 2, 1 \\
		Number of elements / diameter / image & [$-$] & 16 for $d \le$ 25 \textmu m, 20 otherwise  \\
		Number of images / distance to focal plane & [$-$] & 2 \\
		Distance to focal plane & [ \textmu m] & -200, -175, -150, -100, -75, -50, -25 \\
			&  & 0, 25, 50, 75, 100, 125, 150, 200 \\
		\hline
		Total number of elements / diameter & [ $-$ ] & 480 for $d \le$ 25 \textmu m, 600 otherwise \\
		\hline
		\hline
	\end{tabular}
	\label{tab_elem_pool}
\end{table}

\subsection{Results}

\subsubsection{Results from the calibration plate}

\begin{figure}[h]
	\centering
		\includegraphics[width=\textwidth,keepaspectratio]{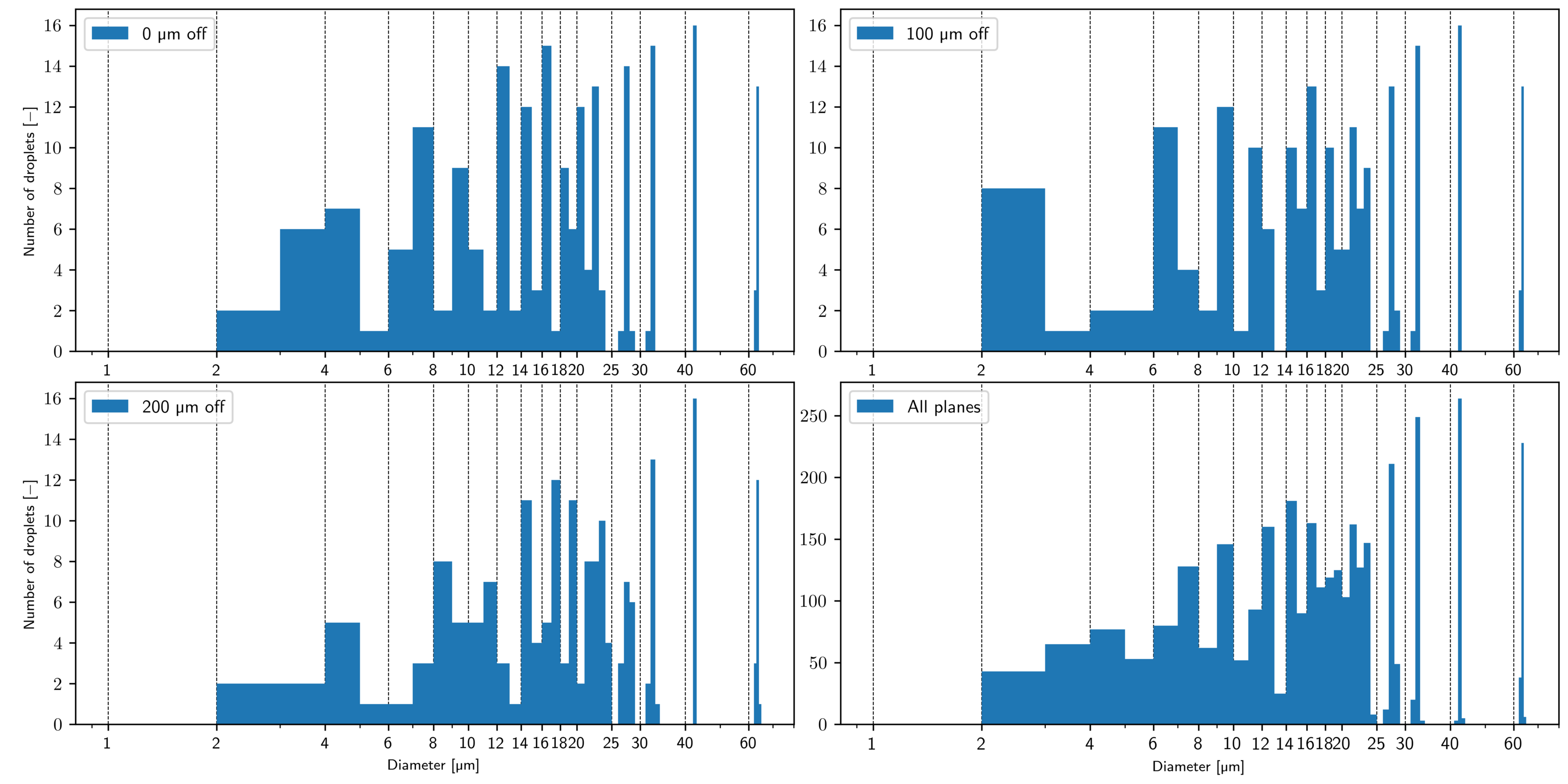}
		\caption{Histograms of the segmented droplet diameter from images of the calibration plate on the focal plane (top left), 100 \textmu m off the focal plane (top right) and 200 \textmu m off the focal plane (bottom left). Bottom right: histogram from all images.}
		\label{fig_qualitative_comparison_UNET_calib_good}
\end{figure}

The present method was applied to the images of the calibration plate, in which each opaque disk appears exactly 16 times per image. On the segmented images a contour recognition algorithm is applied with a threshold value of 0.5 to extract the droplet diameters. Because the segmented image contains only binary values (0 or 1), the threshold has almost no influence on the estimation of the diameters. 
 In Fig.~\ref{fig_qualitative_comparison_UNET_calib_good} the histogram of segmented droplet diameter is given for an off-distance to the focal plane of 0, 100 and 200 \textmu m. The fourth histogram in Fig.~\ref{fig_qualitative_comparison_UNET_calib_good} (bottom right) is obtained by collecting all segmented droplets from all images in one data set. 
The vertical dashed lines correspond to the diameter of the opaque disks, which are equally distributed with 16 samples per image. Ideally, the detection tool should output a patch of 16 droplets on each dashed line.
On all focal planes, large diameters up to 25 \textmu m are correctly segmented, with a slight dispersion on the measured diameter around the expected values.
For droplets smaller than 25 \textmu m on offset images, the dispersion of the diameters increases in a way that the peaks are not distinguishable. However, small droplets are still detected. Finally, all diameters are overestimated by $\approx$2 \textmu m.
The Sauter Mean Diameter for all slices is equal to 43.7 \textmu m, to be compared to the expected one of 42.0 \textmu m. This is consistent with the constant overestimation of 2 \textmu m identified in Fig.~\ref{fig_qualitative_comparison_UNET_calib_good}. This leads to a deviation of 4\% for the SMD.

\subsubsection{Results from test images}

\begin{figure}[h]
	\centering
		\includegraphics[width=0.5\textwidth,keepaspectratio]{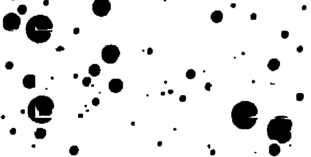}
		\caption{Illustration of the artifact at boundaries where the image is split.}
		\label{fig_UNET_bnd_pb}
\end{figure}

\begin{figure}[h]
	\centering
		\includegraphics[width=0.6\textwidth,keepaspectratio]{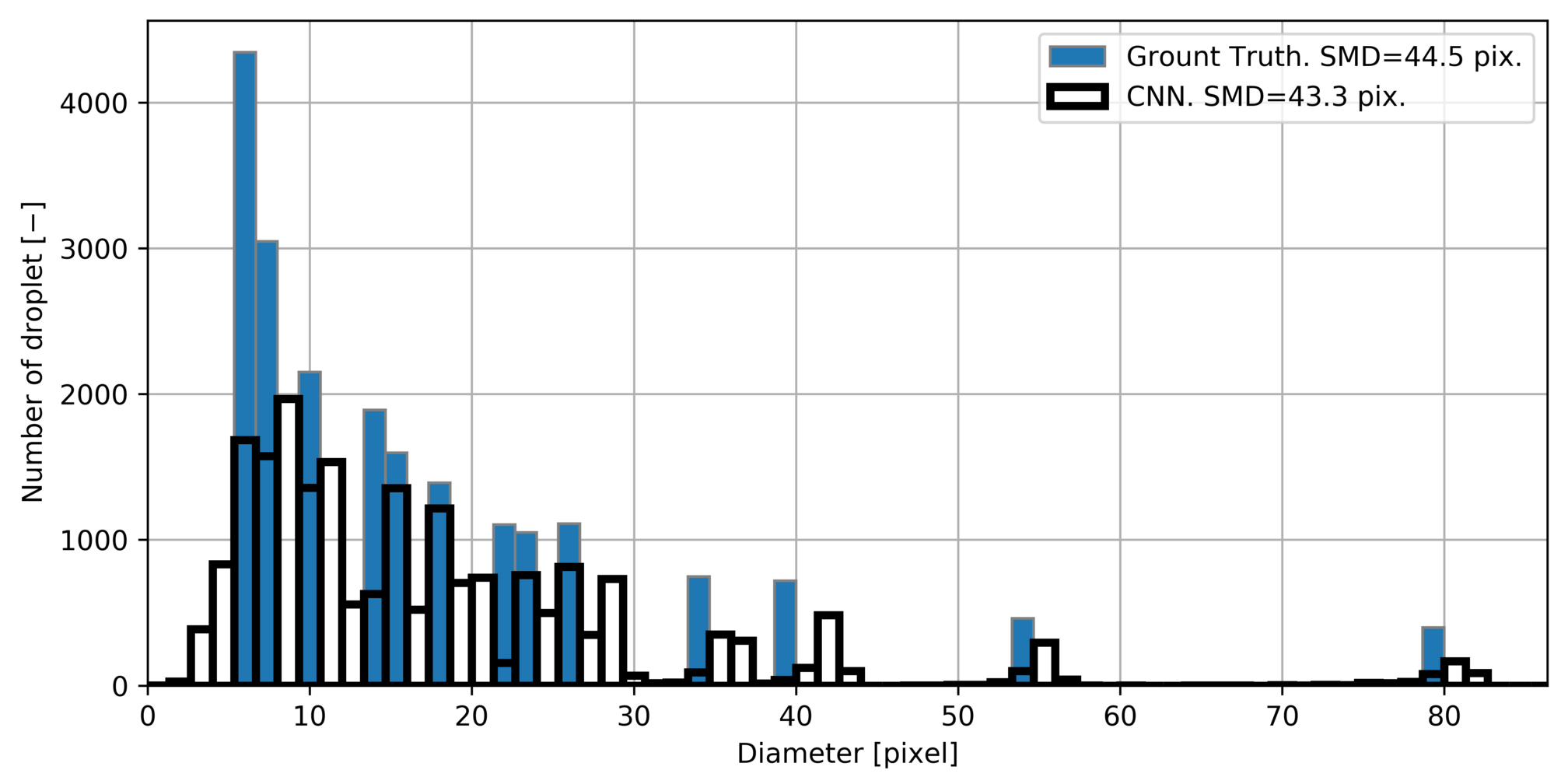}
		\caption{Preliminary result: histogram of segmented droplet diameter from 100 test images}
		\label{fig_UNET_histo_100img}
\end{figure}

The present tool was applied to 100 test images similar the ones depicted in Fig.~\ref{fig_sample_training_data_base}. The results are shown in Fig.~\ref{fig_UNET_histo_100img}. As observed with the calibrations images, large droplets are slightly overestimated. The uncertainty leads to a continuous spectrum for droplets below 10 \textmu m.
The major reason for this discrepancy
is due to the fact that original images are split in smaller images to be post-processed. This leads to a deteriorated post-processing on the boundary of the split image, as depicted in Fig.~\ref{fig_UNET_bnd_pb}. Please note that this artifact is not intrinsic to the principle of the method and will be solved in the future. 
The SMDs are 44.5 and 43.3 pixels for the Ground Truth, and the present method, respectively, leading to a deviation of 2.7\%.

\subsubsection{Results from experimental images in real conditions}
\begin{figure}[h]
	\centering
		\includegraphics[width=0.245\textwidth,keepaspectratio]{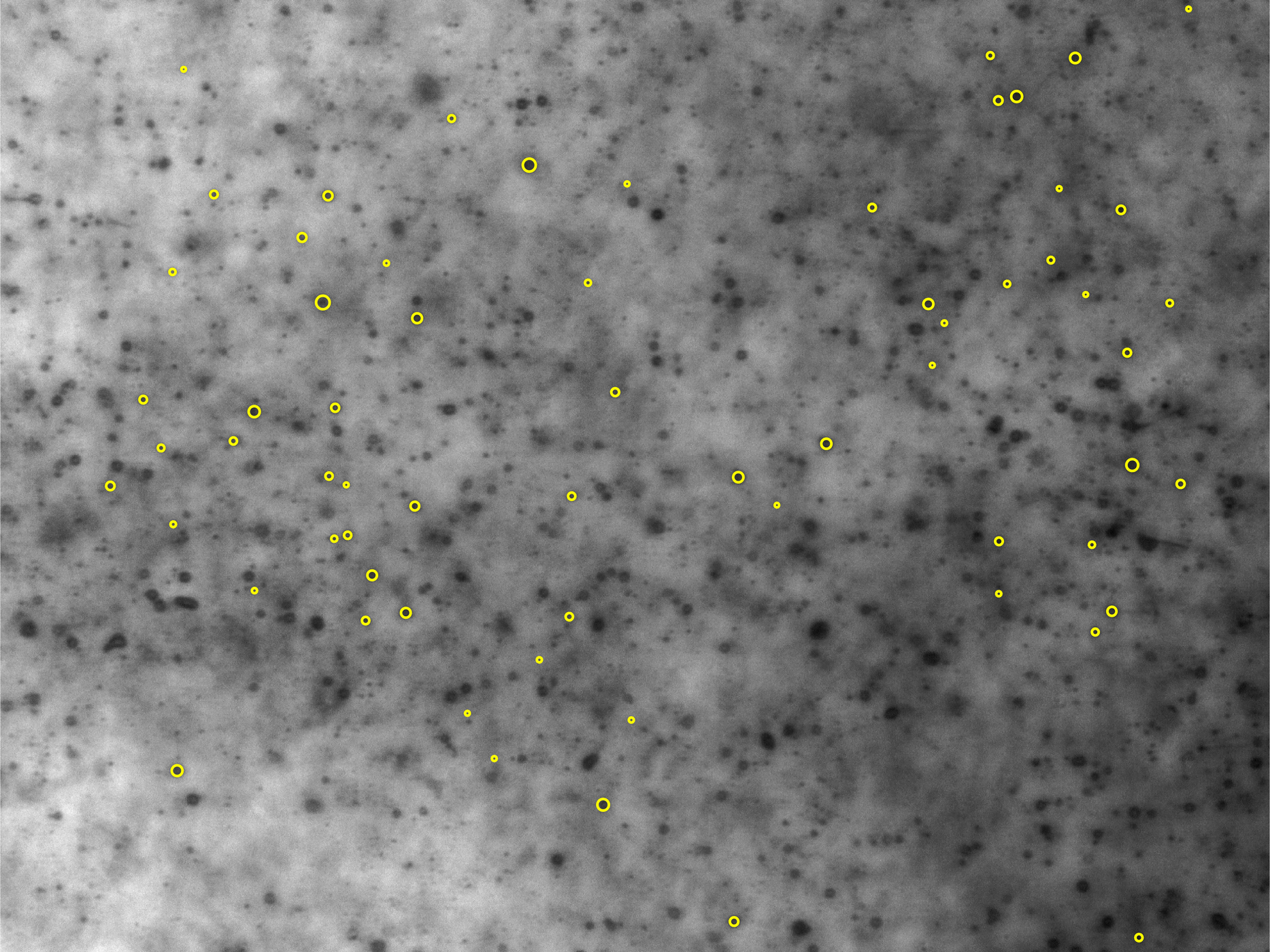}
		\includegraphics[width=0.245\textwidth,keepaspectratio]{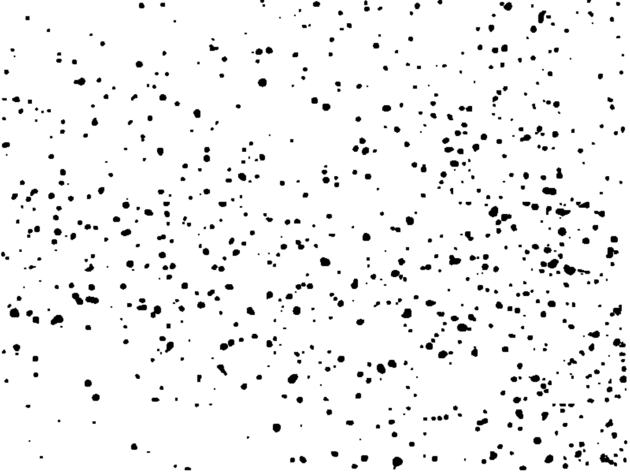}
		\includegraphics[width=0.245\textwidth,keepaspectratio]{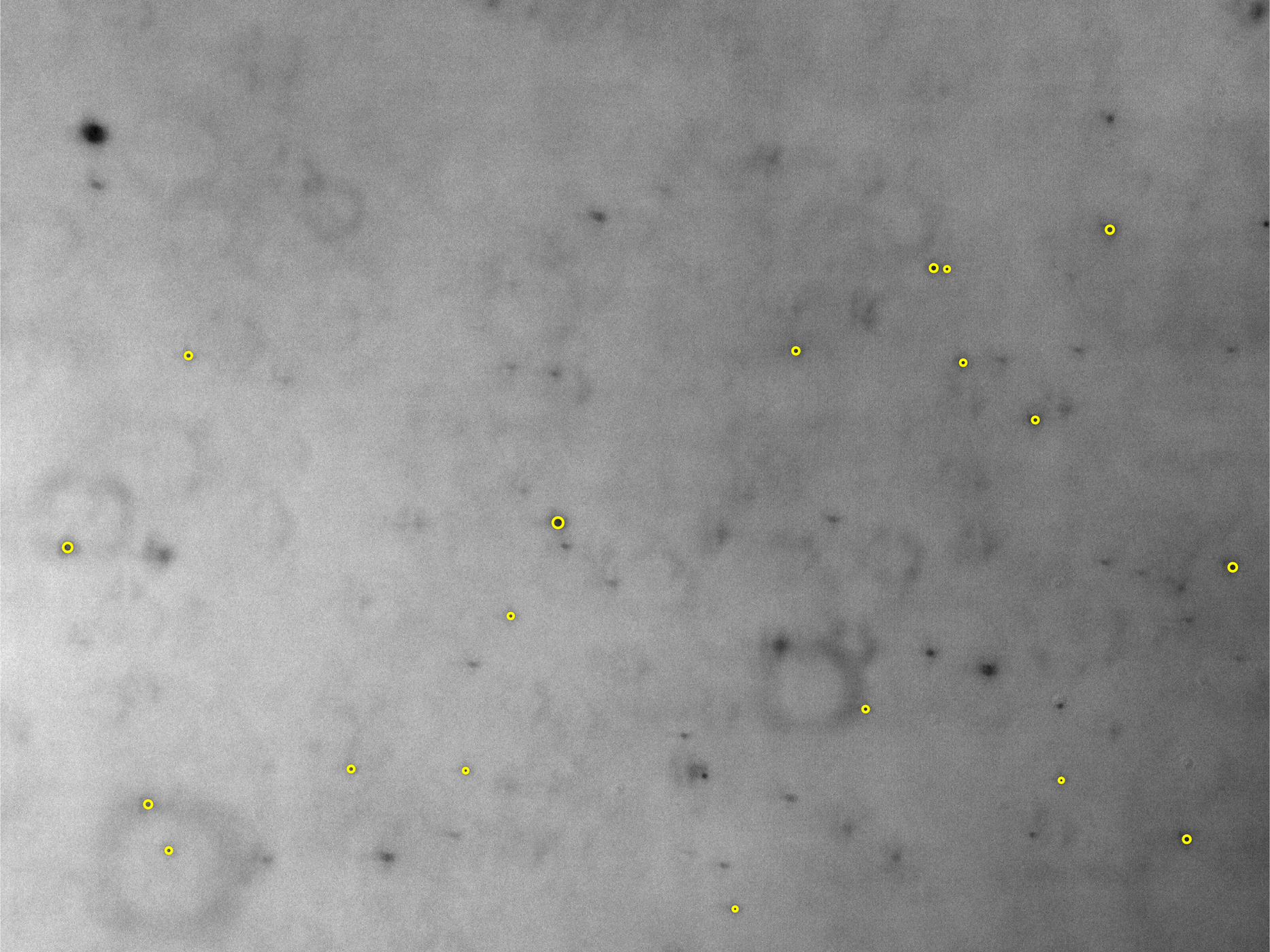}
		\includegraphics[width=0.245\textwidth,keepaspectratio]{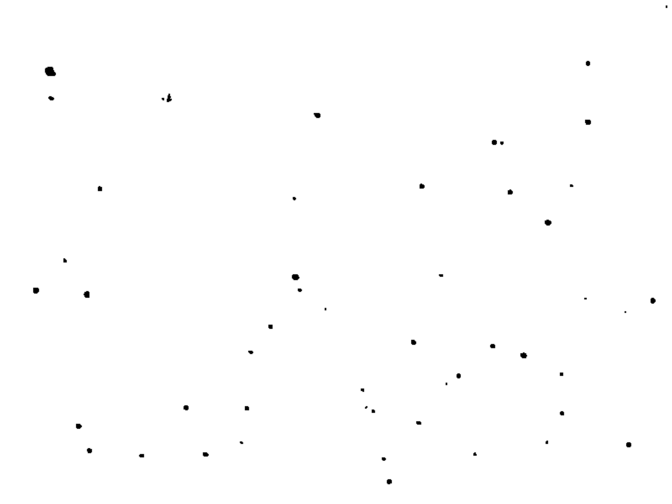}
		\caption{Preliminary result. Left: input image superimposed with typical method (image zone segmentation). Right: output of the present method.}
		\label{fig_qualitative_comparison_UNET}
\end{figure}

\begin{figure}[h]
	\centering

		\includegraphics[width=\textwidth,keepaspectratio]{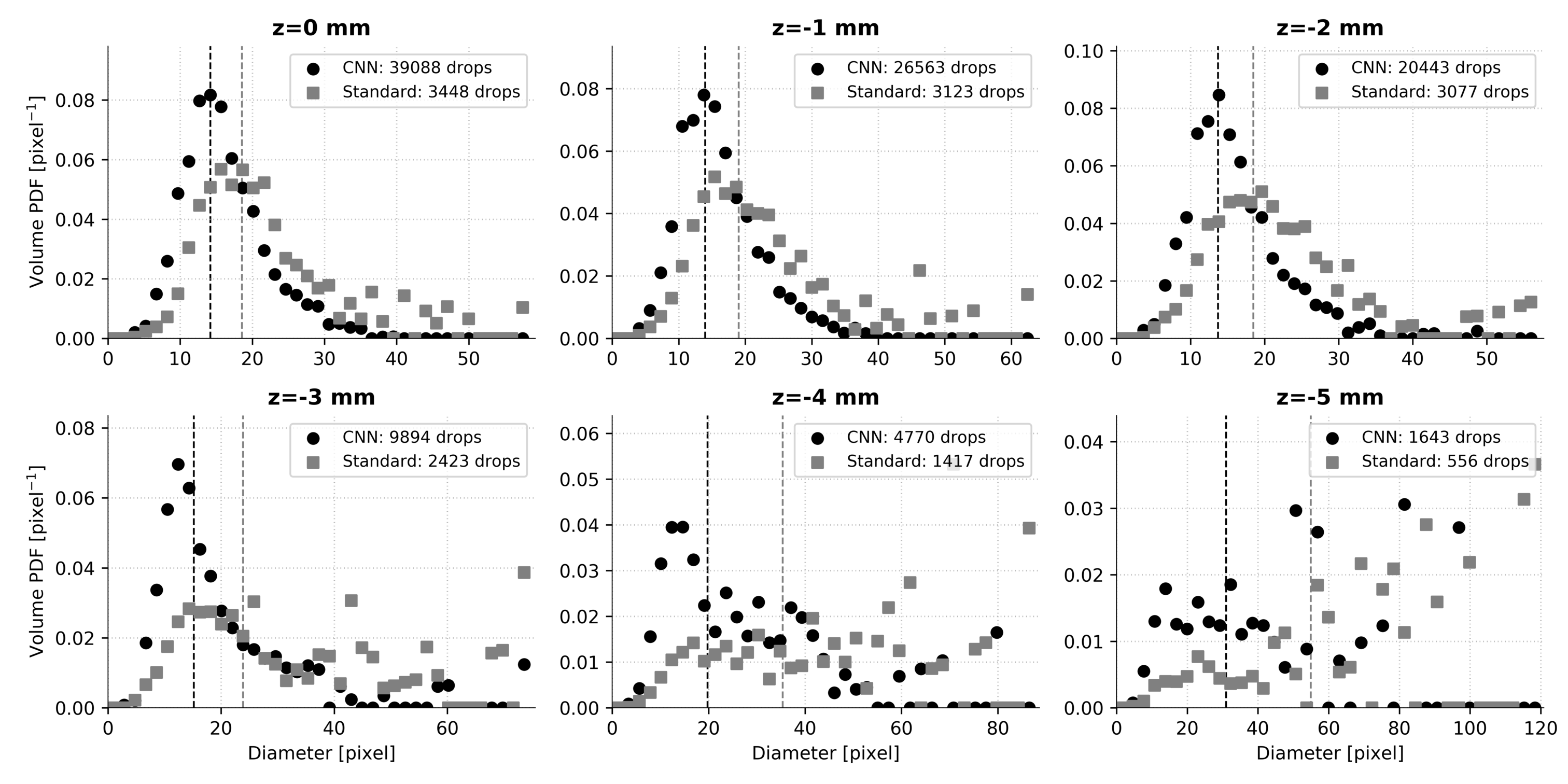}
		\caption{Comparison of volume PDF with the standard and present method.}
		\label{fig_UNET_quant}
\end{figure}

The present post-processing tool was applied to 50 images of the spray in the nozzle vicinity ($x$=50 mm) on each slice, for $z$ ranging from 0 to -5 mm. To assess the benefit of the present method, the same images were post-processed with our standard method. A qualitative comparison of the two methods is given in Fig.~\ref{fig_qualitative_comparison_UNET} at $z$=0 (left) and -5 mm (right). For each slice, two images are disposed side by side. On the first the snapshot is superimposed by the contours detected by the standard method. The second is the segmentation map provided by the present tool.
On the slice $z$=0 mm, the standard method show many false negative (undetected droplets).
This is because of the Depth-of-Field correction \citep{warncke2017experimental} which dismisses droplets whose intensity gradient is not in the range of calibration.
In comparison the proposed method capture much more droplets. The same comments apply at $z$=-5 mm.
The results are presented quantitatively in Fig.~\ref{fig_UNET_quant} where the volume PDF of the droplet diameter is computed separately for each slice. The Sauter Mean Diameter is superimposed as a vertical dashed line.
The number of collected droplets for 50 images is given in the legend for each method. The present method significantly increases the number of collected droplets by a factor ranging from 3 to 11, which allows a better statistical convergence with a restricted number of snapshots. It is observed that the present method estimates a lower SMD than the standard method. The most probable explanation is because (i) much more smaller and (ii) less larger droplets are detected. With the standard method, the tail of the distribution increases with $D^3$ already at $z$=-1 mm, which is symptomatic of a poor statistical convergence. The present method shows this trend later for $z<$-4 mm. For slice from 0 to -3 mm, the distribution is stable. 
To discriminate between the two methods, it is necessary to compare with a Ground Truth image, this will be done in future studies.

\begin{figure}[h]
	\centering
		\includegraphics[width=0.2495\textwidth,keepaspectratio]{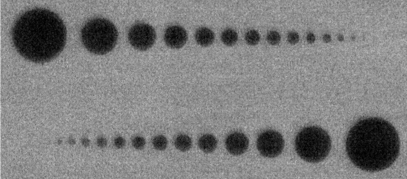}
		\includegraphics[width=0.2495\textwidth,keepaspectratio]{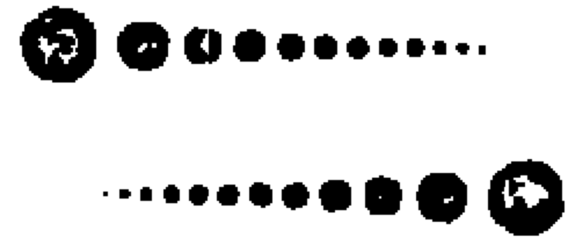} \\
		\includegraphics[width=0.2495\textwidth,keepaspectratio]{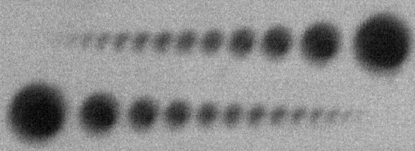}
		\includegraphics[width=0.2495\textwidth,keepaspectratio]{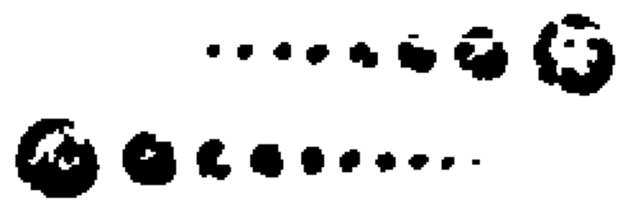}
		\caption{Application of the CNN on new calibration picture with low contrast.}
		\label{fig_unet_calib_low_contrast}
\end{figure}

Finally, additional images of the calibration plate with reduced contrast were post-processed by the present tool.
The reduced contrast was obtained by decreasing the laser intensity when taking the snapshot. Note that this modified contrast is closer to real conditions than the modified contrast obtained from numerical filters in the data augmentation step. The purpose of this test is assess the portability of the CNN to be adapted to configurations different from the ones of the training phase. The results are given qualitatively in Fig.~\ref{fig_unet_calib_low_contrast} where the calibration plate coincides the focal plane (top) and is 100 \textmu m off (bottom). On the focal plane, the proposed method is able to capture small droplets with an acceptable accuracy. The inner part of large droplets is not correctly segmented. Even though this effect is not critical for the detection of small droplets, it stresses the need to train the CNN with real calibration pictures with different light intensity, rather than simulating low contrast with numeric filters.
At 100 \textmu m off the focal plane (Fig.~\ref{fig_unet_calib_low_contrast} bottom), the segmentation error on large droplets is larger, and the small droplets are not detected. However, it is interesting to note that even though intermediate droplets have the shape of an ellipse on the input picture (probably due to a slight angle between the plate and the focal plane), they are correctly segmented as near-to-spherical droplets. This is result is quite promising for further development/refinement of the present method. Also, another possibility to treat deformed droplets is to post-process the segmented image with the computation of the deviation to spherical shape.

\subsection{Conclusion}

The UNET architecture to encode/decode the main feature of an image has shown a promising potential in the post-processing of shadowgraphy images of liquid atomization in degraded conditions. There are still improvements to be done in terms of accuracy. This can be achieved (i) by a better treatment during splitting/merging operations of the final image (boundary artifact), and (ii) by a fine-tuning of the training parameters. 
However, the proposed method allows to capture a considerable larger amount of droplets compared to the traditional method. This is particularly interesting when the number of snapshots is limited (transient process, reduced visibility due to evaporation/obstructed optics). Finally, the CNN show a good capacity to extrapolate a spherical shape based on droplets that where distorted due optical misalignements. To the authors knowledge, this type of correction is out of reach with classical methods.

\section{Detection of overlapping structures and best fitting bounding box \label{sec_overlapping_ellipses}}

The goal of this section is to assess the capability of CNNs to extrapolate the geometrical characteristics of simple shapes with restricted or incomplete information.
This is of relevance for the post-processing of liquid atomization because in the vicinity of the nozzle where the spray is dense, snapshots of the liquid breakup show overlapping structures. Also, X-ray snapshots constitute a superposition of several slices, where the liquid structures strongly overlap.
Traditional methods such as contour detection based on threshold are unable to discriminate two overlapping structures. %
\\In this case we will train a CNN to recognize ellipses that overlap each other. Ellipses are randomly positioned in an image and randomly oriented. The geometrical characteristics to be extracted are the lengths of the major and minor axis as well as their orientation. The information is degraded in two ways.
First, ellipses are randomly disposed in the image, so that they overlap each other and only a portion of their contour is visible.
Second, classical filters as reduced noise, deformation and contrast gradients are added to the picture.

\subsection{Generating the database}

\begin{figure}[h]
	\centering
		\includegraphics[width=0.45\textwidth,keepaspectratio]{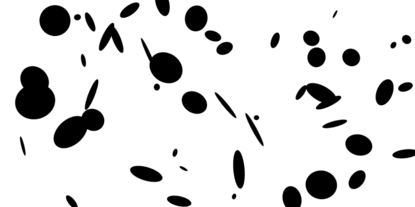}
		\includegraphics[width=0.45\textwidth,keepaspectratio]{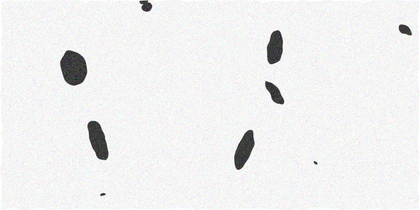}
		\caption{Example of training image with overlap only (left) and typical filters (right).}
		\label{fig_overlapping_ellipse_moderate}
\end{figure}

The different CNN candidates are trained in an increasing difficulty of detection and extrapolation.
First, images contain only a few non-overlapping ellipses, then we increase the number of ellipses on a single image.
Then, we allow ellipses to overlap, and we gradually increase their number. Figure~\ref{fig_overlapping_ellipse_moderate} (left) depicts a moderate number of overlapping ellipses. Then the classical filter for data augmentation are applied to the image (Fig.~\ref{fig_overlapping_ellipse_moderate} right).

\subsection{Selected architectures/models}

To extrapolate the ellipse characteristics from a portion of it, image segmentation and the UNET architecture is not appropriate. Indeed, the area of the image where several ellipses overlap will appear as a marked region, but no information of the related ellipse will be provided. Therefore, instead of segmentation, the task for the CNN is feature detection. The classical output for feature detection is a set of bounding boxes framing the features to detect, usually superimposed to the original image. For this task we can divide the detectors in two categories, depending if they provide a bounding box aligned on the main axes of the image (1\textsuperscript{st} category) or a rotated bounding box that provides a closer fit of the feature to be detected (2\textsuperscript{nd} category). After a literature study, we selected three detectors: SSD by \cite{liu2016ssd}, YOLO by \cite{redmon2016you,redmon2018yolov3} and RRCNN by \cite{jiang2017r2cnn}. They were presented in Sections~\ref{sssec_SSD_pres} to \ref{sssec_RRCNN_pres}.

\subsection{Preliminary results for overlapping Ellipse with YOLO}

\begin{figure}[h]
	\centering
		\hfill
		\includegraphics[width=0.45\textwidth,keepaspectratio]{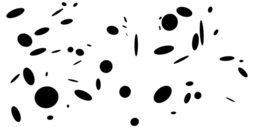}
		\hfill
		\includegraphics[width=0.45\textwidth,keepaspectratio]{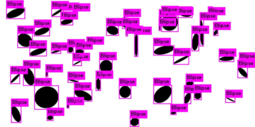}
		\hfill
		\caption{Detection with 50 ellipse without overlapping. 100\% found, Average IOU=0.9.}
		\label{fig_50_nooverlap}
\end{figure}

\begin{figure}[h]
	\centering
		\hfill
		\includegraphics[width=0.45\textwidth,keepaspectratio]{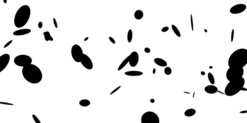}
		\hfill
		\includegraphics[width=0.45\textwidth,keepaspectratio]{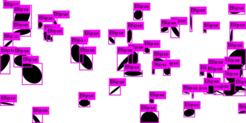}
		\hfill
		\caption{Detection with 50 overlapping ellipse. 98\% found, Average IOU=0.88.}
		\label{fig_50_overlap}
\end{figure}

\begin{figure}[h]
	\centering
		\hfill
		\includegraphics[width=0.45\textwidth,keepaspectratio]{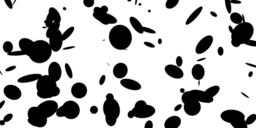}
		\hfill
		\includegraphics[width=0.45\textwidth,keepaspectratio]{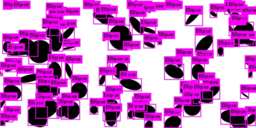}
		\hfill
		\caption{Detection with 100 overlapping ellipse. 89\% found, Average IOU=0.84.}
		\label{fig_100_overlap}
\end{figure}

As a preliminary step to asses the capability of Deep Learning to extrapolate geometrical information, YOLO was tested to detect overlapping ellipses.
During the training, the backpropagation were executed after that 64 images were treated (batch=64). The forward propagation was executed after 4 images (subdivision=16) due to memory constraints. One epoch was made of 1250 images, and the learning rate evolved in the following sequence: 0.001 for step 0-99, then 0.01 for step 100-24999, then 0.001 for step 25000-34999, then 0.0001. The images were resampled to resolution 416x416. 
Results are given in Figs.~\ref{fig_50_nooverlap}-\ref{fig_100_overlap}. First, 50 ellipses without overlapping were randomly inserted (Fig.~\ref{fig_50_nooverlap}) and led to a detection of 100\% and an average IoU score of 0.9. Then 50 overlapping ellipses led to 98\% of detected ellipses with an IoU of 0.88 (Fig.~\ref{fig_50_overlap}). Finally, 100 overlapping ellipses were detected with a success rate of 89\% and an IoU of 0.84 (Fig.~\ref{fig_100_overlap}). 
These preliminary results demonstrated the general capability of Deep Learning for the present purpose. In the rest of this section, the three models are compared.

\subsection{Tests on simple, non-overlapping images}

\begin{figure}[!htb]
	\centering
		\includegraphics[width=0.30\textwidth,keepaspectratio]{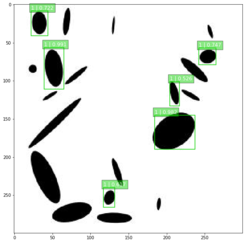}
		\includegraphics[width=0.30\textwidth,keepaspectratio]{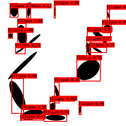}
		\includegraphics[width=0.30\textwidth,keepaspectratio]{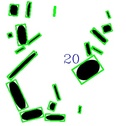}
		\caption{Test of ellipse detection with SSD (left), YOLO (center) and \RRCNN\ (right).}
		\label{fig_BB_comparison}
\end{figure}

In order to asses the capability of detecting simple ellipses, the three models are tested on small images containing 20 ellipses. The three models are trained on the same data base with similar learning parameters. An example of the output is given in Fig.~\ref{fig_BB_comparison}. \RRCNN\ detects all ellipses, YOLO misses a few ellipses of large aspect ratio and SSD performs poorly. It is also observed that the bounding boxes provided by YOLO are sometimes less accurate than the ones by \RRCNN.
The success rate was averaged on 10 images containing 20 ellipses each. SSD, YOLO and \RRCNN\ obtained 53, 90.5 and 99\% success, respectively. Therefore, for the rest of this section, only YOLO and \RRCNN\ will be assessed for further tests. Note that in addition to a better accuracy, \RRCNN\ determines rotated bounding box, which leads to a more precise detection of features.

\subsection{Tests on overlapping ellipses in degraded optical conditions}

\begin{figure}[h]
	\centering
		\includegraphics[width=0.4\textwidth,keepaspectratio]{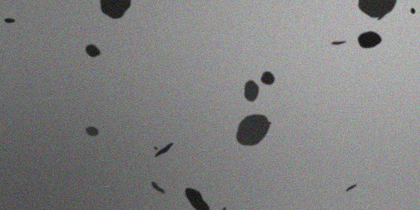}
		\includegraphics[width=0.4\textwidth,keepaspectratio]{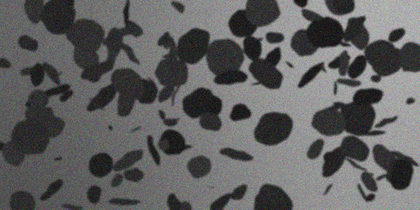}
		\includegraphics[width=0.4\textwidth,keepaspectratio]{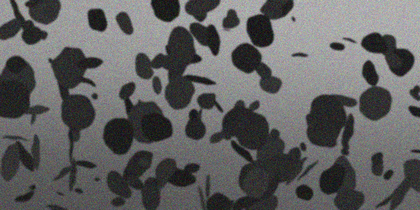}
		\includegraphics[width=0.4\textwidth,keepaspectratio]{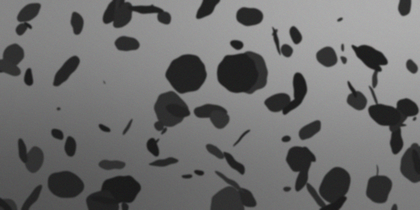}
		\caption{Samples of the training database.}
		\label{fig_ellipse_final_test_database}
\end{figure}

\begin{figure}[h]
	\centering
		\includegraphics[width=0.4\textwidth,keepaspectratio]{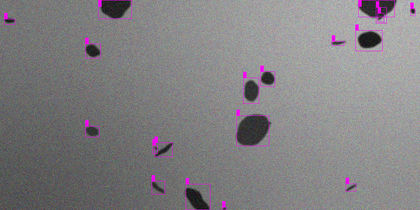}
		\includegraphics[width=0.4\textwidth,keepaspectratio]{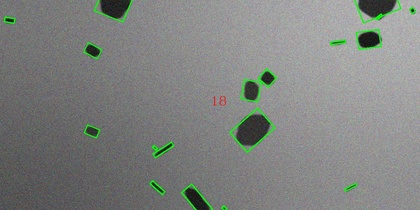}
		\includegraphics[width=0.4\textwidth,keepaspectratio]{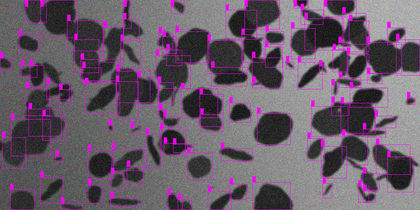}
		\includegraphics[width=0.4\textwidth,keepaspectratio]{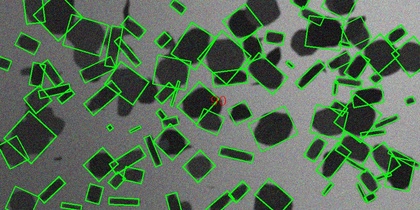}  
		\includegraphics[width=0.4\textwidth,keepaspectratio]{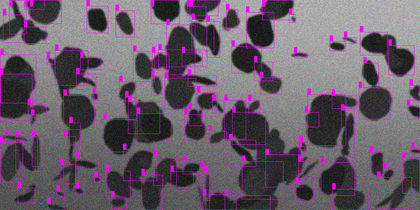}  
		\includegraphics[width=0.4\textwidth,keepaspectratio]{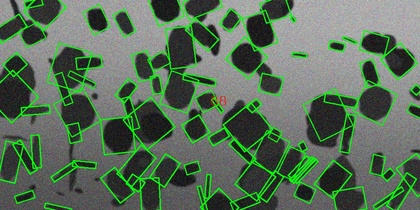}
		\includegraphics[width=0.4\textwidth,keepaspectratio]{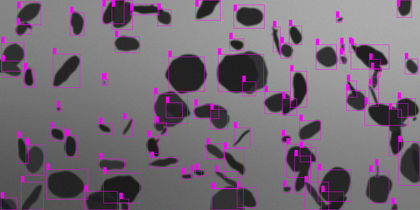}
		\includegraphics[width=0.4\textwidth,keepaspectratio]{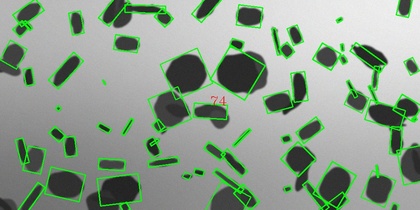}
		\caption{Detected ellipses for YOLO (left) and \RRCNN\ (right).}
		\label{fig_ellipse_final_test_results}
\end{figure}

YOLO and \RRCNN\ are now trained in degraded conditions. The training database consists of 4000 images containing between 20 and 150 ellipses. To mimic degraded conditions, noise, Gaussian blur and deformation are applied to the image. In addition, the contrast between the ellipses and the background is varied, and the background luminosity is made non-homogeneous by applying a constant gradient. Samples of the training images are given in Fig.~\ref{fig_ellipse_final_test_database}.
The models are trained on a similar number of images. YOLO is trained on 192000 images (distributed in 3000 batches of 64 images) while \RRCNN\ is trained on 160000 images.
The predictions for four test images are given in Fig.~\ref{fig_ellipse_final_test_results}. The two models behave very well as they can extrapolate correct bounding boxes for overlapping ellipses, even in case of low contrast, noise, or deformations.\\
The performances indicators presented in Section~\ref{ssec_scores} are estimated on the 360 test images common for YOLO and \RRCNN. They are given in Table~\ref{tab_comp_YOLO_RRCNN}. It is observed that the performance are rather similar. YOLO misses less ellipses, but in counterpart, detects more False Positive than \RRCNN. 

\begin{table}[!htb]
	\centering
	\caption{Performance comparison for YOLO and \RRCNN\ for an IoU threshold of 0.2.}
	\begin{tabular}{l c c c }
		\hline
		\hline
		& Recall & Precision & mAP \\
		YOLO & 0.74 & 0.88 & 0.86 \\
		\RRCNN\ & 0.68 & 0.99 & 0.68 \\
		\hline
		\hline
	\end{tabular}
	\label{tab_comp_YOLO_RRCNN}
\end{table}

\subsection{Conclusion}

Three popular detectors were tested to detect dark ellipses randomly positioned and oriented in a light background. The three networks were trained with the same training parameters. It was found that YOLO and \RRCNN\ provided the best results. In the case of ellipse detection in degraded conditions, YOLO and \RRCNN\ have similar good performances. In addition, one should consider that YOLO provides bounding boxes aligned with the image axis only whereas \RRCNN\ provides rotated bounding box which enclose the feature more closely. With respect to estimating the minor and major axis of ellipse, \RRCNN\ provides instantaneous results with the width and height of the bounding box. In a more general perspective for feature detection in liquid atomization, bounding boxes aligned with the axes of the image are not restrictive. Hence, YOLO is still competitive.
Concerning the capability of extrapolating the geometrical characteristics of overlapping ellipses with Deep Learning, these preliminary tests showed promising results.

\section{Feature detection from experimental image of liquid atomization \label{sec_prefilmer}}

In this section, the feature detector YOLO \citep{redmon2018yolov3} is applied to experimental snapshots of the early spray generated by a planar prefilming airblast atomizer \citep{gepperth2012ligament}. In this type of atomizer, the liquid is disposed on a surface (the prefilmer) in the form a thin film sheared by a high speed air stream. The film is sheared to the tip of the prefilmer where it accumulates in the wake of the prefilmer. Once the liquid accumulation is sufficiently large, it is immersed in the high speed air stream and it fragmented in mainly two types of mechanisms: bag breakup and ligament breakup. The high speed air stream generate an intense turbulence at the tip of the prefilmer. Consequently the surface of the liquid structures (accumulation, ligaments and bags) is highly distorted. Except for droplets and some well-defined long ligaments, this distortion forbids any accurate object detection with traditional methods. 
The goal of this section is to estimate (i) the accuracy of the detector and (ii) the portability of the detector when the CNN is trained only on one type of experiment.

\subsection{Training}

Snapshots from experiment were annotated with bounding box. Different features were annotated:
\begin{enumerate}
\item Attached Ligament: ligaments having an aspect ratio larger than $\approx$3 that are still attached to the liquid accumulation
\item Detached Ligament: in this category we put every structure detached from the liquid accumulation which is not spherical. This includes long ligaments, liquid blobs, but also distorted droplets. This category is quite global and will be split into several sub categories in further studies.
\item Lobe: wavy shape of the liquid accumulation. It can be seen as the precursor of a ligament before it is elongated axially.
\item Bag: the thin liquid membrane created in a bag breakup.
\item Rim: the thick bridge of liquid that frames the membrane in bag breakup.
\end{enumerate}
Note that the round droplets are intentionally kept out of the detected feature, because of the time overhead to manually annotate such numerous structures. Also, in this particular configuration of planar prefilming airblast atomization, most of the droplets are (i) spherical, (ii) distinct from each other and (iii) on a homogeneous background. Therefore, in this configuration, droplets are accurately detected with traditional methods.
In a future study, the capability of YOLO to detect very small objects will be assessed by using a simple contour detector to annotate small droplets.

The training was based on 15 labeled images only, which is a heavy constraint. The gas velocity and the liquid loading were varied between 20 and 60 m/s, and 20 and 120 mm\textsuperscript{2}/s, respectively.
The data was augmented using horizontal and vertical flipping, and modification of exposure, saturation and hue.
During the training, 64 images were treated before backpropagation, the forward propagation being done every two images. Contrary to default parameters, the size of the input was resampled to 672x896 pixel.

\subsection{Results on the same configuration/experiment}

\begin{figure}[h]
	\centering
		\includegraphics[width=0.95\textwidth,keepaspectratio]{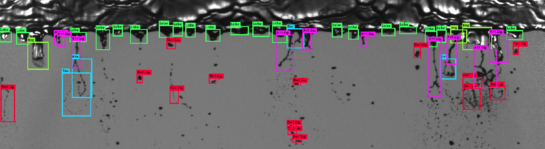}
		\caption{Output of YOLO, gas velocity and film loading are 40 m/s and 20 mm\textsuperscript{2}/s, respectively. Purple: \textit{attached ligament}. Red: \textit{detached ligament}. Dark green: \textit{Lobe}. Light green: \textit{Bag}. Cyan: \textit{Rim}.}
		\label{fig_YOLO_prefilmer01}
\end{figure}

\begin{figure}[h]
	\centering
		\includegraphics[width=0.95\textwidth,keepaspectratio]{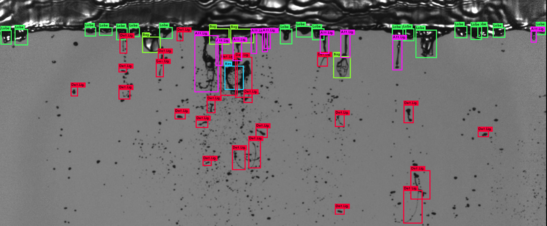}
		\caption{Output of YOLO, gas velocity and film loading are 40 m/s and 20 mm\textsuperscript{2}/s, respectively. Purple: \textit{attached ligament}. Red: \textit{detached ligament}. Dark green: \textit{Lobe}. Light green: \textit{Bag}. Cyan: \textit{Rim}.}
		\label{fig_YOLO_prefilmer02}
\end{figure}

\begin{figure}[h]
	\centering
		\includegraphics[width=0.95\textwidth,keepaspectratio]{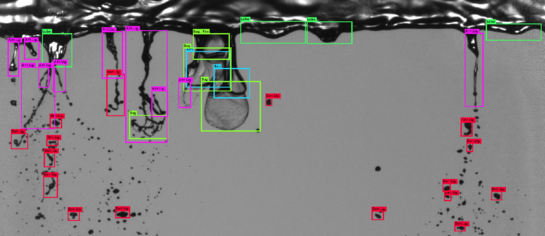}
		\caption{Output of YOLO, gas velocity and film loading are 20 m/s and 20 mm\textsuperscript{2}/s, respectively. Purple: \textit{attached ligament}. Red: \textit{detached ligament}. Dark green: \textit{Lobe}. Light green: \textit{Bag}. Cyan: \textit{Rim}.}
		\label{fig_YOLO_prefilmer03}
\end{figure}

\begin{figure}[h]
	\centering
		\includegraphics[width=0.95\textwidth,keepaspectratio]{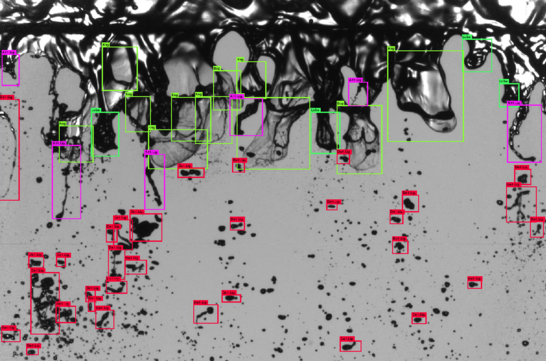}
		\caption{Output of YOLO, gas velocity and film loading are 20 m/s and 120 mm\textsuperscript{2}/s, respectively. Purple: \textit{attached ligament}. Red: \textit{detached ligament}. Dark green: \textit{Lobe}. Light green: \textit{Bag}. Cyan: \textit{Rim}.}
		\label{fig_YOLO_prefilmer04}
\end{figure}

Qualitative results are given in Figs.~\ref{fig_YOLO_prefilmer01} to \ref{fig_YOLO_prefilmer04} for threshold value of 0.2 on the IoU.
Many features are correctly detected, even though they are not all detected (\eg attached or detached ligaments in Fig. \ref{fig_YOLO_prefilmer01}). 
As stated above, small spherical droplets were intentionally left unannotated, which explains why they are not detected in Figs.~\ref{fig_YOLO_prefilmer01} to \ref{fig_YOLO_prefilmer04}.
The performances were quantitatively assessed on 9 images. They are recalled in Table~\ref{tab_assess_YOLO}.
\begin{table}[!htb]
	\centering
	\caption{Performance assessment for feature detection in planar prefilming airblast atomization.}
	\begin{tabular}{c c c c }
		\hline
		\hline
		mean IoU & Recall & Precision & mAP \\
		0.75 & 0.97 & 0.99 & 0.99 \\
		\hline
		\hline
	\end{tabular}
	\label{tab_assess_YOLO}
\end{table}
Given the very small amount of training images, these results are very promising.

\clearpage

\subsection{Results from the same configuration and a different experiment}

The detector was applied to some figures of the article by \cite{braun2019numerical}. In this article, experimental snapshots of the planar prefilmer airblast atomizer are shown and compared to images extracted from highly resolved numerical simulations. 

\begin{figure}[h]
	\centering
		\includegraphics[width=0.80\textwidth,keepaspectratio]{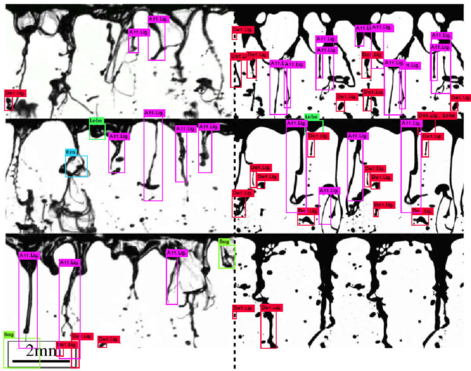}
		\caption{Detector applied to results from experiment (left) and from numerical simulation rendered as in 2D shadowgraphy (right), from \cite{braun2019numerical}}
		\label{fig_YOLO_pref_BRAUN01}
\end{figure}

\begin{figure}[h]
	\centering
		\includegraphics[width=0.80\textwidth,keepaspectratio]{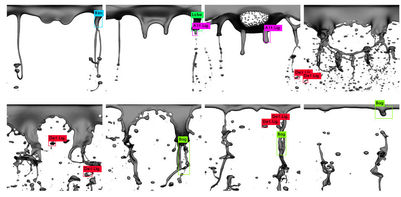}
		\caption{Detector applied to results from numerical simulation rendered in 3D scenery, from \cite{braun2019numerical}}
		\label{fig_YOLO_pref_BRAUN02}
\end{figure}

Figure \ref{fig_YOLO_pref_BRAUN01} shows the output of the detector applied to results from experiment (left) and from numerical simulation rendered as in 2D shadowgraphy (right). The feature detection of the experimental snapshots performs moderately. This might be due to the lower resolution of the figure exported from the article. On the other hand, the results of the numerical simulation have a better detection success rate with almost no false positive.
One can notice that the detector is not repeatable:
the snapshots of the numerical simulation as displayed in Fig.~\ref{fig_YOLO_pref_BRAUN01} are made of a pattern repeated twice. This is possible because periodic boundary conditions were set in the simulation. This means that the features on the image are exactly repeated twice. It is seen that a minority of twin features are detected only once whereas they should be detected twice.
However, this is also a very good result: the detector was trained on 15 images from experiment and Fig.~\ref{fig_YOLO_pref_BRAUN01} proves that it is able to detect the same structures from numerical simulation.
\\Figure \ref{fig_YOLO_pref_BRAUN02} shows the output of the detector applied to results from numerical simulation rendered in 3D scenery. Only a few features are correctly detected. The bags are always false positive. The poor results, compared to Fig.~\ref{fig_YOLO_pref_BRAUN01}, are explained by the fact that the detector was trained on 2D experimental shadowgraphy images, in which the light source is in front of the objective, thus avoiding any perspective effects. In the case of the numerical results of Fig.~\ref{fig_YOLO_pref_BRAUN02}, they were rendered with a light source not aligned with the virtual objective, leading to shades on the liquid surface, which were not learned by the CNN.

\subsection{Results on a different configuration}

\begin{figure}[h]
	\centering
		\includegraphics[width=0.32\textwidth,keepaspectratio]{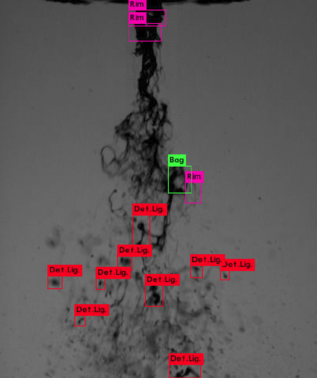}
		\includegraphics[width=0.28\textwidth,keepaspectratio]{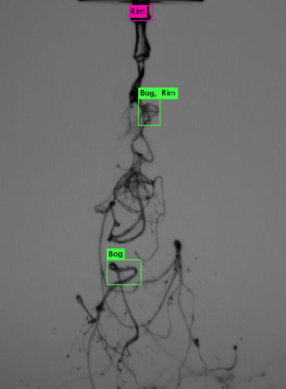}
		\includegraphics[width=0.32\textwidth,keepaspectratio]{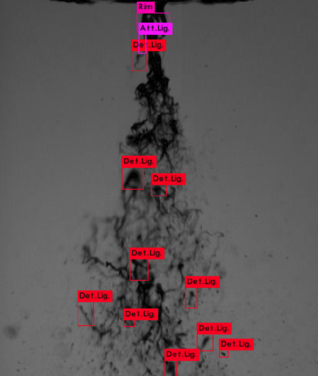}
		\caption{Detector applied to the air-assisted atomization of cylindrical jet, images from \citep{saenger2014viscous,saenger2015experimental}}
		\label{fig_detector_liquid_jet}
\end{figure}

The detector was applied on snapshots from an air-assisted atomization experiment. In this configuration the liquid is injected in the form of a central jet surrounded by a high velocity coflowing airstream \citep{saenger2014viscous}. As in the case of prefilming airblast atomization, the liquid is fragmented in ligaments and bags. One important difference with planar prefilming atomization is the locality of the primary breakup. In planar prefilming atomization, the breakup occurs at the liquid accumulation in the vicinity of the nozzle on a plane that coincides with the focal plane of the camera. In the case of a cylindrical liquid jet, the breakup occurs on the surface of the jet which is (i) convected downstream due to combined effect of the jet velocity and the shearing by the gas, and (ii) subject to flapping and pulsating phenomena. Therefore the position of the ligaments and bags is dispersed inside a volume similar to a cone. This leads to a large amount of features (attached/detached ligaments, bags and rims) out of the focal plane, thus with blurry contours. As the detector was not trained on defocused features, it is expected that few objects will be detected.
The results are shown in Fig.~\ref{fig_detector_liquid_jet}. As expected only few features are detected, mostly detached ligaments and one bag/rim. Nevertheless there are 22 correct features out of 29 total detected features, which is an acceptable results, given (i) the low number of training images and (ii) the different configuration.

\subsection{Conclusion}

With only 15 annotated images and simple data augmentation, the detector based on YOLO succeeded:
\begin{enumerate}
\item to accurately identify many features of the planar prefilming atomization
\item to detect many features from the results of a numerical simulation of the same configuration
\item to detect similar liquid structures generated in another configuration 
\end{enumerate} 
This is a very promising result. In the future, the detector will be trained on more images with a larger spectrum of operating conditions (larger gas velocity, different ambient pressure). Also to develop a more general tool, the detector will trained with images from other configurations, such as air-assisted atomization of a cylindrical jet, jet in cross flow, or pressurized jet as encountered in Internal Combustion Engines.

\section{General conclusion}

The use of CNN for the post-processing of liquid atomization experiments showed a neat improvement for the detection of droplets in degraded optical conditions. The segmentation of the droplets need some further improvement to exploit the full potential of CNNs. The extrapolation of geometrical characteristics of overlapping ellipses in degraded condition showed very promising results with the models YOLO and \RRCNN. Finally the feature detection from experimental snapshots of primary atomization was able to detect most of characteristic structure of primary breakup. It is important to highlight that in this case, the number of training images was very low. This stresses the capability of feature detection tools based on CNN applied on liquid atomization images.
Also, the portability of a detector trained on one experiment to detect liquid structures of another type of experiment was demonstrated. Thus, the potential of Deep Learning for post-processing images of liquid atomization is evident. The next step of this study is to embed these models into \textit{DeepSpray}, a toolbox that relies on CNNs to improve the post-processing of liquid atomization.

\section{Acknowledgement \label{sec_thankyouverymuch}}
The authors like to thank the Helmholtz Association of German Research Centres (HGF) for funding (Grant: 34.14.02).

\section*{References}
{\scriptsize
\bibliography{../../BIBFILES/COMB_STD-jan13,../../BIBFILES/BIB_PERSO,../../BIBFILES/literatur_diss}

\end{document}